\documentclass[12pt,conference]{IEEEtran}
\IEEEoverridecommandlockouts
\usepackage{tikz}
\usepackage{placeins}  
\usepackage{amsmath}
\usepackage{graphicx}
\usepackage{subcaption}
\usepackage{amsfonts}
\usepackage{color, colortbl}
\usepackage{booktabs}
\usetikzlibrary{positioning, shapes, arrows.meta}
\IEEEoverridecommandlockouts
\usepackage{cite}
\graphicspath{ {RoCP/} }

\usepackage{forest}
\usepackage{amsmath,amssymb,amsfonts}
\usepackage{algorithmic}
\usepackage{graphicx}
\usepackage{textcomp}
\usepackage{xcolor}
\usepackage{tikz}
\usepackage{multirow}
\usepackage{hyperref}
\def\BibTeX{{\rm B\kern-.05em{\sc i\kern-.025em b}\kern-.08em
		T\kern-.1667em\lower.7ex\hbox{E}\kern-.125emX}}
\makeatletter
\def\ps@IEEEtitlepagestyle{%
	\def\@oddfoot{\mycopyrightnotice}%
	\def\@evenfoot{}%
}
\def\mycopyrightnotice{%
	{\footnotesize \hfill}
	\gdef\mycopyrightnotice{}
}
\begin{document}
	
	\title{RoCP-GNN: Robust Conformal Prediction for Graph Neural Networks in Node-Classification\\
		{
		}
	}
	
	\author{\IEEEauthorblockN{S. Akansha}
		\IEEEauthorblockA{\textit{Department of Mathematics}\\
			\textit{Manipal Institute of Technology}\\
			Manipal Academy of Higher Education - 576104, India.\\
			akansha.agrawal@manipal.edu.}
			}	
	\maketitle	
\begin{abstract}
Graph Neural Networks (GNNs) have emerged as powerful tools for predicting outcomes in graph-structured data, but their inability to provide robust uncertainty estimates limits their reliability in high-stakes contexts. To address this, we propose Robust Conformal Prediction for GNNs (RoCP-GNN), a novel approach that integrates conformal prediction (CP) directly into the GNN training process to construct statistically robust prediction sets with valid marginal coverage. Applying CP to graph data presents two main challenges: maintaining exchangeability despite interdependencies between nodes, and obtaining smaller, more informative prediction sets. RoCP-GNN addresses both challenges by preserving exchangeability and ensuring high efficiency through smaller, informative prediction sets. RoCP-GNN is model-agnostic and can be integrated with any predictive GNN model, enabling robust predictions and uncertainty quantification in graph-based tasks. Experimental results on standard graph benchmark datasets demonstrate that RoCP-GNN models provide a statistically significant increase in efficiency while maintaining the accuracy of base GNN models across various state-of-the-art architectures for node classification tasks.  The code will be made available upon publication.

	\end{abstract}
	
	
\section{Introduction}	
In recent years, the rapid growth of data across various fields has sparked significant interest in utilizing graph structures to model complex relationships \cite{ran_she_kout-15a, les_mca-12a, def_bre_van-16a, gil_sch_ril-17a, ham_yin_les-17a}. Graphs, comprising nodes that represent entities and edges that capture their interactions, have become a core representation of data in domains such as social networks \cite{les_mca-12a, che_li_bru-17a, min_gao_pen-21a}, recommendation systems \cite{wan_yuy-22a, gao_wan-22a, chu_yao-22a, che_yeh_wan-22a, gao-zhe-li_23a}, drug discovery \cite{bon-bia-sca_21a, han-lak-liu-21a}, fluid dynamics simulation \cite{lin-fot-bha_22a, li-far_22a}, biology \cite{yan_li-23a, jin_eis_son-21a}, and more. The ability of graphs to effectively capture complex interactions between entities makes them an essential tool for analyzing intricate systems. As graph-structured data becomes more diverse and complex, there is an increasing demand for advanced methods to process and interpret these relationships. This has driven the development of Graph Neural Networks (GNNs) \cite{kip_wel-2016a, vel_pet_gui-2017a, ham_yin_les-17a, zhu-21a, gas_etal-19a}, which have proven highly effective across a wide range of tasks in various domains.




As GNNs find increasing deployment in high-stakes settings, understanding the uncertainty inherent in their predictions becomes paramount. The literature presents several methods for uncertainty quantification in Graph Neural Networks (GNNs) for node classification tasks. These include model-agnostic calibration methods \cite{guo-geo-yu_17a,guo-ple-sun_17a,zha-kai-han_20a,gup-rah-aja_20a} and specialized techniques that leverage network principles such as homophily \cite{wan-liu-shi_21a,hsu-she-tom_22a}. Recently, Lin et al. \cite{lin-zha-shi_24a} developed Graph Neural Stochastic Diffusion (GNSD), which integrates Q-Wiener process theory into graph domains to better handle uncertainty in node classification. To address high computational costs, they propose an approximation strategy for discretization sampling of the Q-Wiener process. One prominent approach to uncertainty quantification involves constructing prediction sets that outline a plausible range of values the true outcome may encompass. While numerous methods have been proposed to achieve this objective \cite{hsu-she-tom_22a, zha-kai-han_20a, lak-pri-blu_17a}, many lack rigorous guarantees regarding their validity, specifically the probability that the prediction set covers the true outcome \cite{ang-bat-mal_20a}. This lack of rigor hinders their reliable deployment in situations where errors can have significant consequences.

Conformal prediction (CP) \cite{sha-vov_08a, ang-bat-mal_20a, ang-bat_21a, fon-zen-van_23a} offers a promising framework for constructing statistically valid prediction sets that encompass the true label with user-defined probability, independent of data distribution. CP's reliance on exchangeability—where all permutations of instances are equally likely—makes it particularly suitable for uncertainty quantification in graph-based scenarios, as it relaxes the stringent assumption of independent and identically distributed (i.i.d.) data. Despite CP's success in various machine learning domains, its application to graph data remains underexplored, with most efforts focusing on inductive settings. Notable contributions include Tibshirani et al.'s work on covariate shift \cite{tib-foy-can_19a}, Gibbs and Candes' adaptive CP for online settings with time-varying data distributions \cite{gib-can_21a, gib-can_22a}, Plachy, Makur, and Rubinstein's approach to label shift in federated learning \cite{pla-mak-rub_23a}, and Akansha's method for obtaining confidence intervals under conditional data shift \cite{aka_24a}. 

Current research in the graph domain primarily employs Conformal Prediction (CP) either as a wrapper mechanism for determining prediction sets \cite{zar-ant-boj_23a, zar-boj_24a} or as a means to design effective post-hoc calibration functions for pre-trained Graph Neural Network (GNN) classifiers, enhancing prediction efficiency \cite{hua-jin-can_24a}. Our work builds upon and extends the research of Huang et al. \cite{hua-jin-can_24a}, who proposed a correction model leveraging node dependencies and a calibration set to adjust base GNN predictions, ensuring prediction intervals meet desired coverage probabilities. Their approach computes non-conformity scores for the calibration set to adjust future predictions, using a GNN model to correct the base GNN's prediction probabilities, thereby improving efficiency. In contrast, our work introduces a novel end-to-end training setup that offers significant advantages. By integrating uncertainty quantification directly into the GNN training process, our method eliminates the need for separate correction models, enhancing computational efficiency. Crucially, this integrated approach simultaneously improves both prediction accuracy and the reliability of confidence intervals, potentially leading to more robust and accurate uncertainty estimates in GNN-based node classification tasks without the need for additional post-processing steps.   

Our main contributions can be summarized as follows:
\begin{enumerate}
	\item \textbf{Integration of Conformal Prediction during Training}: We incorporate conformal prediction (CP) directly into the training process, moving beyond traditional post-hoc methods. We introduce a novel size-loss metric for GNN classifiers, designed to optimize the generation of compact prediction sets. This enhancement leads to improved computational efficiency and interpretability.
	
	
	
	
	\item \textbf{Empirical Validation on Diverse Graph Datasets}: We incorporate conformal prediction (CP) directly into the training process, moving beyond traditional post-hoc methods. We introduce a novel size-loss metric for GNN classifiers, designed to optimize the generation of compact prediction sets. This enhancement leads to improved computational efficiency and interpretability.\footnote{For detailed information on accuracy improvements, please refer to the experiments and results section.}.
	\item \textbf{Model Agnostic and Implementation Simplicity}: Our method is model-agnostic, offering seamless integration with any graph node classifier model. Its straightforward implementation makes it accessible to a broad spectrum of researchers and practitioners, facilitating widespread adoption and fostering collaborative advancements in uncertainty-aware machine learning frameworks.
\end{enumerate}

\section{Background and problem framework} \label{sec:background} 
This section outlines the problem formulation addressed in this article. Additionally, we provide relevant background information essential for applying conformal prediction (CP) as a wrapper to achieve marginal coverage at test time.
\subsection{Problem settings} \label{subsec:ProbSet}
This study addresses the task of node classification within a graph under transductive settings. We consider an undirected graph \( G = (V, E, X, A) \), where \( |V| = n \) represents the number of nodes, and \( E \subseteq V \times V \) denotes the set of edges. Each node \( v_i \in V \) is associated with a feature vector \( x_i \in \mathbb{R}^d \), and the input feature matrix is given by \( X \in \mathbb{R}^{n \times d} \). The adjacency matrix \( A \in \{0,1\}^{n \times n} \) is defined such that \( A_{ij} = 1 \) if \( (v_i, v_j) \in E \), and \( A_{ij} = 0 \) otherwise. Additionally, each node \( v_i \) is assigned a label \( y_i \in Y = \{1, \ldots, K\} \), where \( y_i \) represents the ground-truth label. The dataset is denoted as \( D := (X, Y) \), which is initially divided into training, calibration, and test sets. The training set is further split into training and validation subsets, denoted as \( D_{\text{train}} \), \( D_{\text{valid}} \), \( D_{\text{calib}} \), and \( D_{\text{test}} \), with fixed sizes. During the training process, the GNN has access to the data \( \{(x_v, y_v)\}_{v \in D_{\text{train}} \cup D_{\text{valid}}} \), the attribute information \( \{x_v\}_{v \in D_{\text{calib}} \cup D_{\text{test}}} \), and the full graph structure \( (V, E, A) \). However, the labels \( \{y_v\}_{v \in D_{\text{calib}} \cup D_{\text{test}}} \) remain hidden throughout the training phase. 

\subsection{Graph Neural Networks in Node Classification}
Graph Neural Networks (GNNs) are designed to learn compact representations that integrate both the structural information of the network and the attributes of the nodes. This learning process is facilitated through a series of propagation layers \cite{gil_sch_ril-17a}, represented by the equation:
{\small
	\begin{align}\label{eq:gnnlayer}
		h^{(l)}_u & = UP_{(l-1)}\biggl\{h^{(l-1)}_u, \vspace{1cm} AGG_{(l-1)}\{h^{(l-1)}_v \mid v \in N_u\}\biggr\}
\end{align}} 
In this equation, \( h^{(l-1)}_u \) denotes the node representation at the \((l-1)\)-th layer, initially set to the node's feature vector at the first layer, and \( N_u \) represents the set of neighbors for node \( u \). Each layer performs two key operations: Aggregation and Update. The Aggregation function, \( AGG_{(l-1)}(\cdot) \), collects information from neighboring nodes at the \((l-1)\)-th layer and is defined as \(AGG_{(l-1)} : \mathbb{R}^{d_{(l-1)}} \times \mathbb{R}^{d_{(l-1)}} \to \mathbb{R}^{d_{(l-1)}'}\). The Update function, \( UP_{(l-1)}(\cdot) \), combines the aggregated information into the node's representation at the ((l-1))-th layer and is defined as \(UP_{(l-1)} : \mathbb{R}^{d_{(l-1)}} \times \mathbb{R}^{d_{(l-1)}'} \to \mathbb{R}^{d_{(l)}}\). Through iterative application of this message-passing mechanism, GNNs refine node representations by considering their neighbors' relationships.

In classification tasks, given an input node feature vector \( x_v \), objective is to estimate the posterior distribution over a set of classes \( \{1, 2, \ldots, K\} \), represented by \( \mathbf{p}_\theta(x_v) \) for each node \( v \). Here, \( \theta \) denotes the parameters of the classifier model \(\pi_{\theta}(x_v)\), which in our case is a Graph Neural Network (GNN). The predicted probability that node \( v \) belongs to class \( j \) is denoted as \( {p}_{\theta,j}(x_v) \), where \( j = 1, 2, \ldots, K \), and is given by \( {p}_{\theta,j}(x_v) = P(Y = y_j \mid X = x_v) \). Bayes’ decision rule is then applied to choose the class with the highest posterior probability, thereby minimizing the classification loss. This approach allows the model \( \pi_{\theta}(x_v) \) to achieve high accuracy on test datasets. However, high accuracy alone is insufficient to guarantee safe and reliable deployment in practice. 
\subsection{Conformal Prediction}	Conformal Prediction (CP) \cite{vov-gam-sha_05a} is a statistical methodology designed to generate prediction sets that encompass the true outcome with a user-defined level of confidence. Mathematically, CP generates confidence sets \( C(x) \subseteq Y = {1, 2, \ldots, K} \), ensuring that the true class \( y \) is included in the confidence set \( C(x) \) with a probability of at least \( 1 - \epsilon \), where \( \epsilon \in (0,1) \) is a user-defined error rate. More formally, this can be expressed as \( P(y \in C(x)) \geq 1 - \epsilon \), under the assumption that the calibration examples \( (x_v, y_v) \), \( v \in {D}_{\text{calib}} \), are exchangeably sampled from the test distribution.

In this work, for computational efficacy we focus solely on the split conformal approach\footnote{Although the "full conformal" approach can yield more precise results, it is often computationally prohibitive.}, , also known as the inductive conformal method \cite{pap-pro-vov_02a}, for computational efficacy. The CP process can be broken down into three key stages, given a predetermined error rate $\epsilon$:

\noindent1. \textit{Defining Non-conformity Measures:} The first step involves establishing a heuristic metric, known as the non-conformity score $S: X \times Y \rightarrow \mathbb{R}$. This score quantifies how well a label $y$ aligns with the prediction for input $x$. In a classification context, this might be represented by the predicted probability of a particular class $y$.	

\noindent2. \textit{Calculating the Quantile:} The next phase involves determining the $(1-\epsilon)$-quantile of the non-conformity scores of the calibration dataset $D_\text{calib}$. This is expressed as:	
\( \tilde{\eta} = \text{quantile}(\{S(x_1, y_1), \ldots, S(x_p, y_p)\}, (1-\epsilon)(1 + \frac{1}{p})), \)
where $p = |D_\text{calib}|,$ the number of samples in \(D_{calib}\).	
	
\noindent3. \textit{Constructing the Prediction Set:} For a new test sample $x_{p+1}$, we form a prediction set:	
	\[ C(x_{p+1};\tilde{\eta}) = \{y \in Y : S(x_{p+1}, y) \leq \tilde{\eta}\} \]
Assuming the exchangeability of $\{(z_i)\}_{i=1}^{p+1} = \{(x_i, y_i)\}_{i=1}^{p+1}$, it follows that $S_{p+1} := S(x_{p+1}, y_{p+1})$ is exchangeable with $\{S_i\}_{i=1}^p$. Consequently, $C(x_{p+1})$ encompasses the true label with the specified coverage probability \cite{vov-gam-sha_05a}: \(P\{y_{p+1} \in C(x_{p+1};\tilde{\eta})\} = P\{S_{p+1} \geq \text{quantile}(\{S_i\}_{i=1}^{p+1}, 1 - \epsilon) \geq 1 - \epsilon \).
It's worth noting that this framework is versatile and can accommodate various non-conformity scores.

\subsection{Adaptive Prediction Set (APS)} For test time coverage we use adaptive prediction set (APS) as non-conformity score proposed in \cite{rom-ses-can_20a} specifically designed for classification tasks. This score calculates the cumulative sum of ordered class probabilities until the true class is encountered. Formally, we denote the cumulative probability up to the \( k \)-th most promising class as \( S_\theta(x, k) = \sum_{j=1}^{k} p_{\theta, \pi(j)}(x) \), where \( \pi \) is a permutation of \( Y \) such that \( p_{\theta, \pi(1)}(x) \geq p_{\theta,\pi(2)}(x) \geq \ldots \geq p_{\theta, \pi(K)}(x) \). Subsequently, the prediction set is constructed as \( C(x;\tilde{\eta}) = \{\pi(1), \ldots, \pi(k^*)\} \), where \( k^* = \inf\{ k : \sum_{j=1}^{k} p_{\theta, \pi(j)}(x) \geq \tilde{\eta} \} \).


\subsection{Evaluation metrics} To evaluate the effectiveness of conformal prediction as a wrapper method, we employ two key metrics \cite{hua-jin-can_24a,aka_24a}. These metrics are essential for assessing the validity of marginal coverage and the degree of inefficiency. The first metric, termed 'Coverage', quantifies the empirical marginal coverage across the test dataset \( D_{\text{test}} \). It is calculated as follows:
\begin{equation}
	\text{Coverage} := \frac{1}{|D_{\text{test}}|} \sum_{i \in D_{\text{test}}} I(Y_i \in C(X_i))
\end{equation}
The second metric, termed  'Ineff', measures inefficiency by examining the cardinality of the prediction set. It is calculated as follows:
\begin{equation}\label{eq:ineff}
	\text{Ineff} := \frac{1}{|D_{\text{test}}|} \sum_{i \in D_{\text{test}}} |C(X_i)|
\end{equation}
It's worth noting that a higher value of 'Ineff' indicates greater inefficiency, as it suggests larger prediction sets. It's crucial to understand that this measure of inefficiency in conformal prediction is separate from the accuracy of the underlying model's original predictions.

\subsection{Conformal Prediction in Graph Neural Networks}\label{sec:validCP}
The application of Conformal Prediction (CP) for uncertainty quantification in Graph Neural Networks (GNNs) has recently gained traction, particularly in the context of node classification within transductive settings. Huang et al. \cite{hua-jin-can_24a} demonstrated the feasibility of this approach, highlighting a crucial property of GNNs that enables the use of CP: permutation invariance. In the GNN training process, both calibration and test information are utilized. However, a key observation is that the model's output and non-conformity scores remain consistent regardless of node ordering. This permutation invariant condition is fundamental, as it allows for the valid application of CP to GNN models. Essentially, it ensures that the statistical guarantees provided by CP hold true in the graph-structured data context, where relationships between nodes are inherently important. Building upon these findings, Gazin et al. \cite{gaz-bla-roq_24a} have expanded the theoretical framework, deriving generalized results for transductive settings in regression tasks. This extension broadens the applicability of CP beyond classification problems.

\begin{figure*}[ht]
	\centerline{\includegraphics[height=7.5cm,width=16cm]{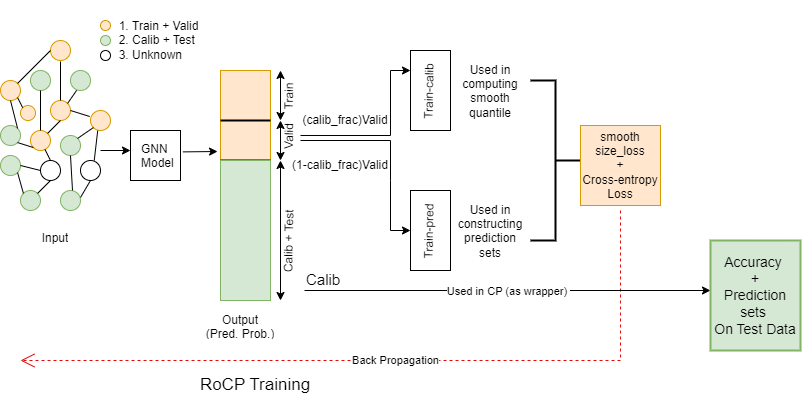}}
	\caption{\textbf{RoCP training illustration: } This illustration demonstrates the key components of Robust conformal Prediction (RoCP) training. }
	\label{fig:rocp_archit}
\end{figure*}
\section{RoCP-GNN: Robust Conformal Prediction for Graph Neural Networks} Traditional GNN classifiers are trained to optimize cross-entropy loss, without considering the efficiency of post-hoc conformal prediction steps. Our proposed Robust Conformal Prediction (RoCP) framework introduces an efficiency-aware size loss, integrated with cross-entropy loss, on which the final GNN classifier is trained. This approach ensures good prediction accuracy alongside efficient prediction sets in downstream steps. 

\subsection{RoCP Implementation }
The implementation of RoCP, as illustrated in Figure \ref{fig:rocp_archit}, involves several key steps in each epoch:

\noindent1. \textbf{Data Partitioning}: We leverage the validation set \(D_{valid}\) by allocating a fraction calib\_frac to \(D_{train-calib}\) for calibration, with the remainder serving as \(D_{train-pred}\) for prediction and loss computation.

\noindent2. \textbf{Smooth Quantile Computation}: Given a miscoverage \(\epsilon\), we compute a differentiable quantile \(\tilde{\eta}\) as the \(\epsilon(1+\frac{1}{|D_{train-calib}|})\)-quantile of the conformity scores \(S_\theta(x,k):= p_{\theta,j}(x)\). This computation utilizes smooth sorting approaches \cite{blo-teb-ber_20a, stu-cem-dou_21a,pet-bor-kue_21a} which typically include a "dispersion" hyper-parameter $\delta$. As $\delta \rightarrow 0$, smooth sorting approximates "hard" sorting. This ensures that the threshold $\tilde{\eta}$ is differentiable with respect to the calibration examples' predictions and the model parameters.

\noindent 3. \textbf{Smooth Prediction Set Construction}: We employ the threshold conformal predictor (THR) \cite{sad-lei-was_19a} to construct prediction sets \(C_{\theta}(x,\tilde{\eta}):={k:p_{\theta,j}(x) =S_\theta(x,k)\ge \tilde{\eta}}\). The subscript \(\theta\) indicates the dependence of confidence sets on the model parameters. We smooth the thresholding of conformity scores using the sigmoid function and a temperature parameter \(T\) \cite{stu-cem-dou_21a}:
\begin{equation}
	C_{\theta,k}(x; \tilde{\eta}) := \sigma \left( \frac{S_{\theta}(x,k) - \tilde{\eta}}{T} \right).
\end{equation}
Here, $C_{\theta,k}(x; \tilde{\eta}) \in [0, 1]$ represents a soft assignment of class $k$ to the confidence set. As $T \rightarrow 0$, we recover the 'hard' confidence set.

\noindent4. \textbf{Loss Function} We define the loss function on which our Graph Neural Network (GNN) is trained as $$\mathcal{L}_{\text{RoCP-GNN}} := \mathcal{L}_{\text{cross-entropy}} + \lambda \mathcal{L}_{\text{size-loss}}.$$ Here, $\mathcal{L}_{\text{cross-entropy}} = \frac{1}{|D_{\text{train}}|}\sum_{v \in D_{\text{train}}}l(y_v,\hat{y}_v)$, where $l$ is the individual loss on each instance in the training sample and $\hat{y}_v$ is the predicted label through the model prediction $\mathbf{p}_\theta(x_v)$ for $v \in D_{\text{train}}$. The size loss is defined as $$\mathcal{L}_{\text{size-loss}} := \frac{1}{|D_{\text{train-pred}}|}\sum_{v \in D_{\text{train-pred}}} C_\theta(x_v),$$ where $C_\theta(x_v) := \frac{1}{K}\sum_{k=1}^{K}(C_{\theta,k}(x_v;\tilde{\eta})-\tau)$. The parameter $\lambda$ is the size weight, and by default, we use $\tau = 1$ to avoid penalizing singletons.

It is important to note that RoCP training optimizes the model parameters $\theta$, on which the confidence sets $C_{\theta}$ depend through the model predictions $\mathbf{p}_{\theta}$. The size loss $\mathcal{L}_{\text{size-loss}}$ is a "smooth" loss intended to minimize the expected inefficiency, i.e., $\mathbb{E}[|C_{\theta}(X; \tau)|]$, and should not be confused with the statistic used for evaluation as post-hoc. $C_{\theta,k}(x; \tilde{\eta})$ can be interpreted as a soft assignment of class $k$ to the confidence set $C_{\theta}(x; \tilde{\eta})$.

After training, any Conformal Prediction (CP) method can be applied to re-calibrate $\tilde{\eta}$ on a held-out calibration set $D_{\text{calib}}$ as usual. The thresholds $\tilde{\eta}$ obtained during training are not retained, ensuring that we obtain the coverage guarantee of CP. While these smooth operations might seem to compromise the coverage guarantee, we recover the original non-smooth computations and the corresponding coverage guarantee in the limit of $T, \delta \rightarrow 0$. In practice, we empirically obtain coverage close to $(1 - \epsilon)$, which is sufficient for training purposes. At test time, we revert to the original (non-smooth) implementations, ensuring that the coverage guarantee holds as per established literature.

	\begin{table}[ht]
		\centering
		\caption{Dataset Statistics}\label{tab:dataset}
\begin{tabular}{llll}
	\hline\\ [-0.7em]
	\multirow{2}{*}{} & \multirow{2}{*}{} & \multirow{2}{*}{} &\multirow{2}{*}{}  \\
	\cline{1-4}\\[-0.7em]
	&Cora & Citeseer & Pubmed \\
	\hline \\[-0.7em] \vspace{.1cm}
	\#nodes & 2708 & 3327 & 19717 \\ \vspace{.1cm}
	\#edges& 10556 & 9104 & 88648 \\ \vspace{.1cm}
	\#features & 1433  & 3703 & 500 \\ \vspace{.1cm}
	\#classes& 7  & 6 & 3 \\ \vspace{.1cm}
	$H(G)$& 0.81 & 0.74 & 0.80 \\ \vspace{.1cm}
	Directed  & $\times$ & $\times$ & $\times$ \\ 
	\hline
\end{tabular}
	\end{table}
\begin{table*}[ht]
	\centering
	\caption{\textbf{Main Inefficiency Results.} This table presents the empirical inefficiency for node classification, measured by the size of the prediction set (smaller numbers indicate better efficiency). We compare RoCP training (denoted as RoCP) with standard cross-entropy training (denoted as CP). The relative improvement (in \%) of RoCP over standard training is indicated on the arrows. Results show the average and standard deviation of prediction set sizes, calculated from 10 GNN runs, each with 200 calibration/test splits. The miscoverage rate $\epsilon$ is set to 0.1 (top table) and 0.05 (bottom table).}
	\label{table:efficiency}
	\begin{tabular}{l|l|l|l}
		\hline \\[-0.7em]
		Model & $~~~~~~~~~~~~~~~~~$Cora & $~~~~~~~~~~~~~~~~~$ Citeseer & $~~~~~~~~~~~~~~~$Pubmed \\\hline \\[-0.7em] \vspace{.1cm}
		& $~~~~~$CP $\longrightarrow $ RoCP & $~~~~~$CP $\longrightarrow $ RoCP & $~~~~~$CP $\longrightarrow $ RoCP \\ \hline \\[-0.7em] \vspace{.1cm}
		APPNP & $3.6\pm 1.6 \xrightarrow{-52.22\%} 1.72\pm 0.90$ & $2.8 \pm 0.8 \xrightarrow{-29.29\%} 1.98 \pm 0.42$ & $1.8 \pm 0.42 \xrightarrow{-13.89\%} 1.55 \pm 0.57$ \\ \vspace{.1cm}
		GCN & $3.4\pm 1.1 \xrightarrow{-45.88\%} 1.84\pm 0.23$  & $2.9 \pm 0.2 \xrightarrow{-25.86\%} 2.15 \pm 1.28$ & $1.6 \pm 0.32 \xrightarrow{-2.62\%} 1.57 \pm 1.05$  \\ \vspace{.1cm}
		GAT & $3.0\pm 0.9 \xrightarrow{-42.67\%} 1.72 \pm 0.99$  & $2.6 \pm 0.6 \xrightarrow{-22.69\%} 2.01 \pm 1.12$ & $2.1 \pm 1.32 \xrightarrow{-20.48\%} 1.67 \pm 0.95$  \\ \vspace{.1cm}
		GraphSAGE & $2.8\pm 1.9 \xrightarrow{-28.21\%} 2.01 \pm 1.06$  & $2.8 \pm 1.1 \xrightarrow{-32.14\%} 1.9 \pm 1.25$ & $1.7 \pm 1.56 \xrightarrow{-9.41\%} 1.54 \pm 1.25$ \\
		\hline \\[-0.7em] \vspace{.1cm}
		APPNP & $4.4 \pm 1.6 \xrightarrow{-44.77\%} 2.43 \pm 1.13$  & $3.6 \pm 0.9 \xrightarrow{-30.28\%} 2.71 \pm 1.8$ & $2.6 \pm 0.4 \xrightarrow{-20.64\%} 2.4 \pm 1.43$ \\ \vspace{.1cm}
		GCN & $4.2 \pm 1.4 \xrightarrow{-37.62\%} 2.62 \pm 1.69$  & $3.6 \pm 0.3 \xrightarrow{-5.00\%} 3.42 \pm 1.12$ & $2.61 \pm 0.4 \xrightarrow{-0.38\%} 2.6 \pm 1.29$ \\ \vspace{.1cm}
		GAT & $3.7 \pm 1.2 \xrightarrow{-31.08\%} 2.55 \pm 1.32$  & $3.5 \pm 0.9 \xrightarrow{-10.29\%} 3.14 \pm 1.31$ & $2.67 \pm 0.4 \xrightarrow{-17.98\%} 2.19 \pm 0.57$ \\ \vspace{.1cm}
		GraphSAGE & $3.5 \pm 1.9 \xrightarrow{-16.29\%} 2.93 \pm 1.83$  & $3.8 \pm 1.7 \xrightarrow{-10.53\%} 3.40 \pm 1.14$ & $2.8 \pm 0.4 \xrightarrow{-19.29\%} 2.26 \pm 1.16$ \\
		\hline
	\end{tabular}
\end{table*}
	
\section{Experiments and Results} 
This section is divided into two main parts, both focusing on node classification tasks using Graph Neural Networks (GNNs). First, we validate that our RoCP training approach maintains the accuracy of the base GNN models for node classification. Second, we compare RoCP training with standard cross-entropy training in terms of reducing the inefficiency of the post-training calibration technique APS, using the metric 'Ineff' defined in Equation~\eqref{eq:ineff}. The main results with optimized parameters are presented in Tables~\ref{table:accuracy} and~\ref{table:efficiency}, respectively. We consider several benchmark node classification graph datasets, with their statistics provided in Table~\ref{tab:dataset}, as well as different GNN architectures. We report metrics averaged across 100 random calibration/test splits for 10 trained models for each method. Our focus is on the (non-differentiable) APS as a Conformal Prediction method used after training, which allows us to obtain the corresponding coverage guarantee for node classification. This approach enables us to evaluate the effectiveness of our RoCP training method compared to standard cross-entropy training, specifically in reducing the inefficiency of APS in the context of classifying nodes within graph-structured data using GNNs.


\subsection{Datasets, Models, and Evaluation}

Our experimental analysis focuses on the semi-supervised node classification task using three well-known benchmark datasets: Cora \cite{mcc_nig_ren-00a}, CiteSeer \cite{sen_nam-08a}, and PubMed \cite{nam_lon-12a}. These datasets represent paper citation networks where node features are bag-of-words extracted from paper content, and the objective is to classify research topics. Table \ref{tab:dataset} presents the statistics of these datasets, including the graph homophily $H(G)$, defined in \cite{pei_wei_cha-20a} as:

\begin{equation}\label{eq:graphhomophily}
	H(G) = \frac{1}{|V|}\sum_{v\in V}\frac{\text{$v$'s neighbors with the same label as $v$}}{N_v}
\end{equation}
For the calibration data $D_{\text{calib}}$, we utilize $\min \{1000, (|D_{\text{calib}} \cup D_{\text{test}}|)/2\}$ samples, with the remaining samples designated as test data. We assume that the calibration and test data samples are drawn from the same distribution. Each model is trained for 200 epochs. During RoCP training, we use a fraction (calib\_frac) of $D_{\text{valid}}$ for computing the smooth quantile $\tilde{\eta}$, and the remainder of $D_{\text{valid}}$ for the prediction step. The impact of the parameter calib\_frac is examined in the ablation study presented in Section \ref{sec:ablation}. Our main results correspond to optimized parameters.

To evaluate the effectiveness of our approach across different GNN architectures, we consider two categories of state-of-the-art graph node-classification models. The first category includes models that incorporate integrated message passing and transformation operations, such as GCN \cite{kip_wel-2016a}, GAT \cite{vel_pet_gui-2017a}, and GraphSage \cite{ham_wil_jur-2017a}. The second category comprises models that treat message passing and nonlinear transformation as separate operations, including APPNP \cite{gas_etal-19a} and DAGNN \cite{liu_gao_ji-20a}. This diverse selection of models enables a comprehensive assessment of our RoCP training approach across various GNN architectures.

\subsection{Inefficiency Results} In this section, we focus on comparing the inefficiency reduction achieved by RoCP training (denoted as RoCP in the main result table) to that of a standard cross-entropy training baseline (denoted as CP in the main result table). Table \ref{table:efficiency} presents the main results on inefficiency for $\epsilon = 0.1$ (top table) and $\epsilon = 0.2$ (bottom table). As evident from the main results, RoCP consistently reduces inefficiency across all datasets and state-of-the-art GNN models. Our approach is novel in incorporating Conformal Prediction (CP) directly into the GNN training pipeline. Consequently, we compare the results obtained using our RoCP approach solely with standard cross-entropy training. While our RoCP training utilizes the THR conformal predictor \cite{sad-lei-was_19a}, it is worth noting that any conformal predictor could potentially be used in place of THR during training. Alternative methods include: Adaptive Prediction Sets (APS) \cite{rom-ses-can_20a}, Regularized Adaptive Prediction Sets (RAPS) \cite{ang-bat-mal_20a,ang-bat_21a},  Defused Adaptive Prediction Sets (DAPS) \cite{zar-ant-boj_23a}. This flexibility in choice of conformal predictor further enhances the potential applicability of our RoCP training approach across various scenarios and requirements.


\subsection{RoCP Maintains Accuracy}
The main accuracy results are presented in Table \ref{table:accuracy}. As evident from these results, the RoCP approach consistently maintains the accuracy of the base GNN models across almost all datasets. Notably, in many cases, RoCP training actually improves the accuracy by 1-2\%. This improvement demonstrates that RoCP training is not only cost-efficient but also preserves and potentially enhances the accuracy of the base model while simultaneously reducing inefficiency. In our main results, we compare the accuracy obtained using RoCP training with that of standard cross-entropy training. The results from both training methods are comparable across all datasets and GNN architectures, further validating the effectiveness of our approach. This consistency in performance across various models and datasets underscores the robustness and versatility of the RoCP training method in maintaining model accuracy while providing the additional benefits of conformal prediction.

\begin{figure}[ht]
	\centerline{\includegraphics[height=5.5cm,width=8cm]{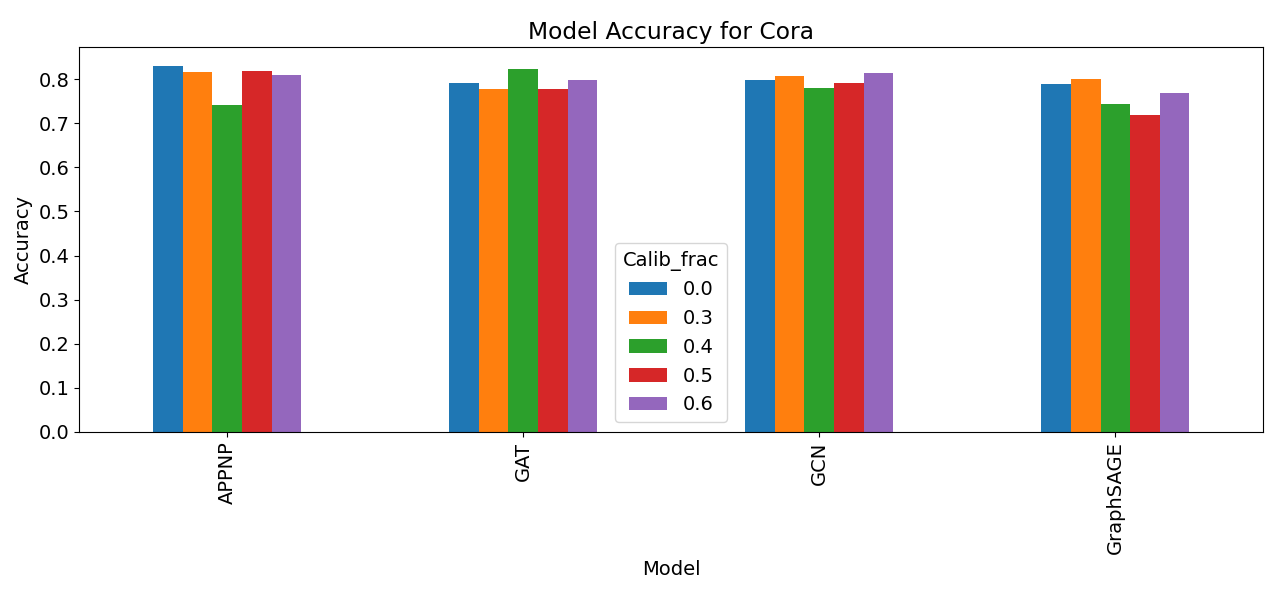}}
	\centerline{\includegraphics[height=5.5cm,width=8cm]{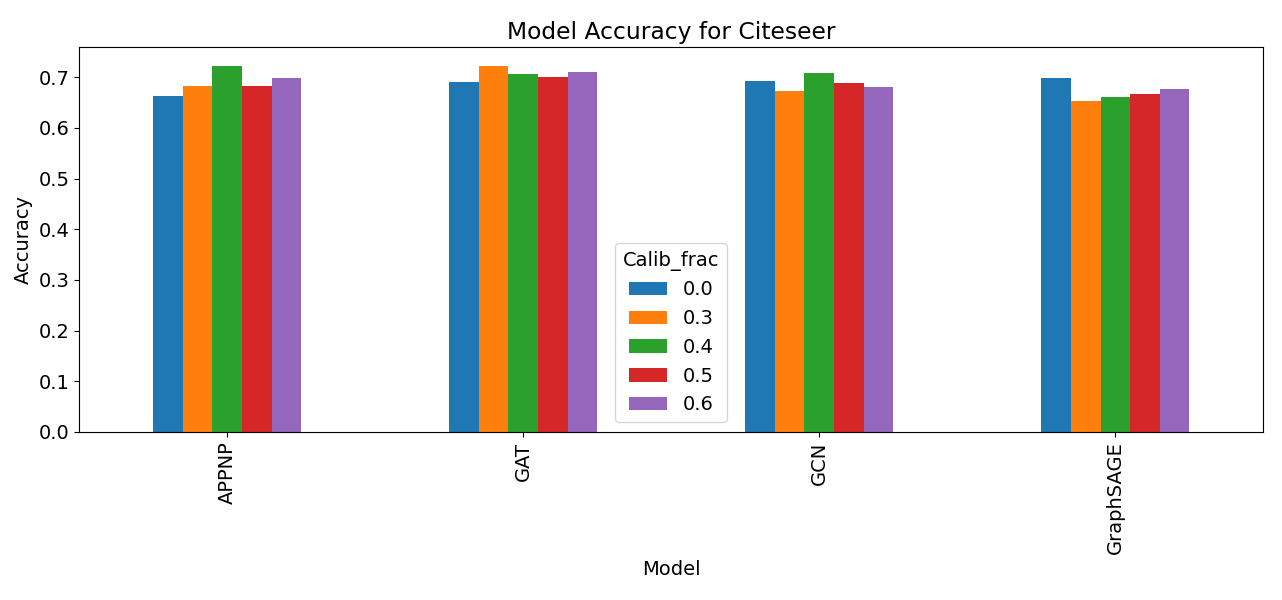}}
	\caption{These bar plots depict the effect of the calibration dataset size, $D_{train-calib}$, on the accuracy of classifier models trained under RoCP, corresponding to calib\_frac values of $\{0.3, 0.4, 0.5, 0.6\}$. The results are compared to standard cross-entropy training, represented by calib\_frac = 0.0, across various GNN models.}
	\label{fig:accuracy_calib}
\end{figure}
\begin{figure}[ht]
	\centerline{\includegraphics[height=5.5cm,width=8cm]{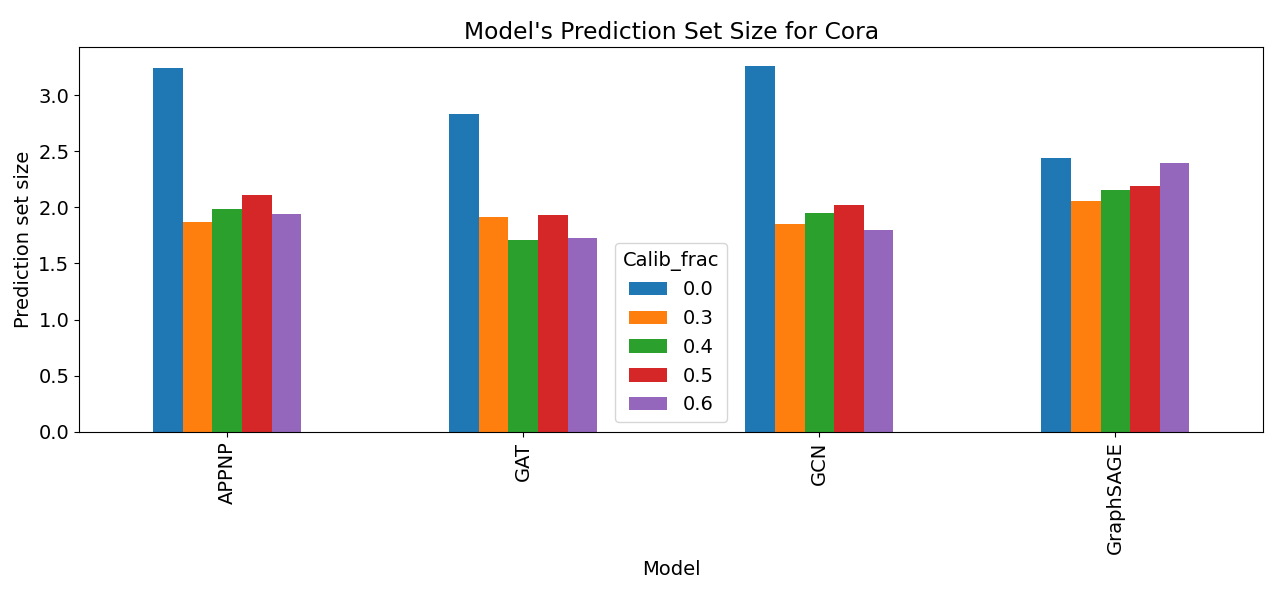}}
	\centerline{\includegraphics[height=5.5cm,width=8cm]{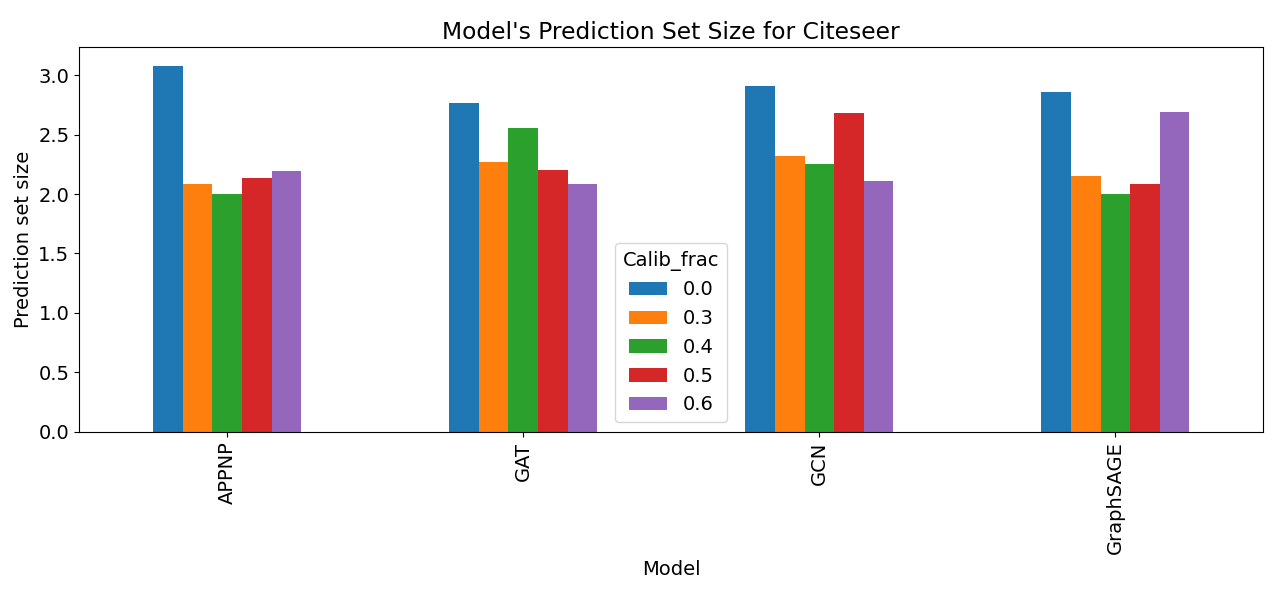}}
	\caption{These bar plots illustrate the cumulative impact of the calibration dataset size, $D_{train-calib}$, on the prediction set size for different models trained with RoCP, corresponding to calib\_frac values of $\{0.3, 0.4, 0.5, 0.6\}$. The results are compared to models using only CP, represented by calib\_frac = 0.0, across various GNN models.}
	\label{fig:predsize_calib}
\end{figure}

\begin{table*}[ht]
	\centering
	\caption{\textbf{Main Accuracy Results.} This table presents the accuracy results of four state-of-the-art (SOTA) GNN models on citation network benchmarks, comparing standard cross-entropy training (CP) with our RoCP approach. Accuracy is measured using $F_1$-micro scores. The transition from CP to RoCP is shown in the columns, with relative changes (in \%) indicated on the arrows. These results demonstrate that our approach maintains, and in most cases improves, the accuracy of the base models. Values represent the average and standard deviation of $F_1$-micro scores, based on 10 independent GNN runs.}
	\label{table:accuracy}
\begin{tabular}{l|l|l|l}
	\hline \\[-0.7em]
	Model & $~~~~~~~~~~~~~~~~~$Cora &$~~~~~~~~~~~~~~~~~$ Citeseer & $~~~~~~~~~~~~~~~$Pubmed \\\hline \\[-0.7em] \vspace{.1cm}
	&  $~~~~~$CP $\longrightarrow $ RoCP & $~~~~~$CP $\longrightarrow $ RoCP & $~~~~~$CP $\longrightarrow $ RoCP \\ \hline \\[-0.7em] \vspace{.1cm}
	APPNP & $83.11\pm 0.7 \xrightarrow{-0.24\%}$ \text{82.91 $\pm$ 0.70 } & $69.28 \pm 0.8 \xrightarrow{5.17\%}$ \text{72.86 $\pm$ 0.61} &  $79.51\pm 2.2 \xrightarrow{0.25\%} 79.71 \pm 2.9$ \\ \vspace{.1cm}
	GCN & $80.56\pm 0.4 \xrightarrow{1.19\%} 81.52\pm 0.57$ & $70.02 \pm 0.4 \xrightarrow{1.79\%} 71.27 \pm 0.73$ &  $76.09\pm 1.5 \xrightarrow{1.39\%} 77.15 \pm 1.7$ \\ \vspace{.1cm}
	GAT & $81.09\pm 0.2 \xrightarrow{0.55\%} 81.54 \pm 0.86$ & $69.68 \pm 0.4 \xrightarrow{1.35\%} 70.62 \pm 0.51$ &  $77.39\pm 1.9 \xrightarrow{0.17\%} 77.52 \pm 1.5$ \\ \vspace{.1cm}
	GraphSAGE & $79.59\pm 0.9 \xrightarrow{0.63\%} 80.09 \pm 0.34$ & $68.19 \pm 0.62 \xrightarrow{1.32\%} 69.09 \pm 1.7$ &  $75.91\pm 2.5 \xrightarrow{0.32\%} 76.15 \pm 0.91$ \\
	\hline
\end{tabular}

	
	
\end{table*}

\begin{table*}[ht]
	\centering
	\caption{\textbf{Coverage Results.} This table presents the empirical marginal coverage for node classification tasks. Results are shown for miscoverage rates $\epsilon = 0.1$ (upper table) and $\epsilon = 0.05$ (lower table). Values represent the average and standard deviation of coverage, calculated from 8 independent GNN runs, each with 200 calibration/test splits. The coverage is computed using Conformal Prediction (CP) applied post-training.}
	\label{table:coverage}
\begin{tabular}{l|l|l}
	\hline \\[-0.7em]
	Model & $~~~~~~~~~~~~~~~~~$Cora & $~~~~~~~~~~~~~~~~~$ Citeseer \\ \hline \\[-0.7em] \vspace{.1cm}
	APPNP & $90.10 \pm .00 \xrightarrow{-0.08} 90.07 \pm .00$ & $90.10 \pm .00 \xrightarrow{+0.03} 90.25 \pm .00$ \\ \vspace{.1cm}
	DAGNN & $90.13 \pm .00 \xrightarrow{-0.08} 90.02 \pm .00$ & $90.18 \pm .00 \xrightarrow{-0.08} 90.19 \pm .00$ \\  \vspace{.1cm}
	GCN & $90.06 \pm .00 \xrightarrow{+0.03} 90.22 \pm .00$ & $90.07 \pm .00 \xrightarrow{-0.01} 90.23 \pm .00$ \\  \vspace{.1cm}
	GAT & $90.24 \pm .00 \xrightarrow{+0.01} 90.14 \pm .00$ & $90.10 \pm .00 \xrightarrow{+0.02} 90.17 \pm .00$ \\  \vspace{.1cm}
	GraphSAGE & $90.24 \pm .00 \xrightarrow{-0.04} 90.02 \pm .00$ & $90.04 \pm .00 \xrightarrow{+0.08} 90.20 \pm .00$ \\ \hline \\[-0.7em] \vspace{.1cm}
	APPNP & $95.05 \pm .00 \xrightarrow{-0.01} 95.07 \pm .00$ & $95.05 \pm .00 \xrightarrow{+0.04} 95.21 \pm .00$ \\  \vspace{.1cm}
	DAGNN & $95.05 \pm .00 \xrightarrow{-0.09} 95.05 \pm .00$ & $95.12 \pm .00 \xrightarrow{-0.05} 95.22 \pm .00$ \\  \vspace{.1cm}
	GCN & $95.05 \pm .00 \xrightarrow{-0.01} 95.14 \pm .00$ & $95.04 \pm .00 \xrightarrow{-0.05} 95.22 \pm .00$ \\ \vspace{.1cm}
	GAT & $95.20 \pm .00 \xrightarrow{-0.11} 95.04 \pm .00$ & $95.07 \pm .00 \xrightarrow{+0.01} 95.18 \pm .00$ \\  \vspace{.1cm}
	GraphSAGE & $95.19 \pm .00 \xrightarrow{+0.03} 95.08 \pm .00$ & $95.02 \pm .00 \xrightarrow{-0.01} 95.12 \pm .00$ \\ \hline
\end{tabular}
\end{table*}
\subsection{Coverage} Table \ref{table:coverage} presents the coverage results for five different GNN models, aiming at 90\% ($\epsilon = 0.1$) and 95\% ($\epsilon = 0.05$) coverage levels on two benchmark datasets for citation graph node classification. These results confirm that marginal coverage is maintained during RoCP training in transductive settings, supporting the theoretical claims discussed in Section \ref{sec:validCP}. Importantly, RoCP consistently achieves marginal coverage across all datasets and GNN baselines while also improving efficiency by reducing the size of the prediction sets and enhancing accuracy.

\subsection{Ablation}\label{sec:ablation}
Our approach incorporates two main hyperparameters: $\lambda$, the penalty parameter for size-loss, and calib\_frac, which determines the impact of calibration ($D_{\text{train-calib}}$) and prediction ($D_{\text{train-pred}}$) sets during RoCP training on both accuracy and inefficiency. This ablation study examines the influence of these parameters on model performance.

We investigate the impact of varying sample sizes in $D_{\text{train-calib}}$ and $D_{\text{train-pred}}$ on the accuracy and inefficiency of the post-calibration technique. We employ calib\_frac values of $\{0.0, 0.3, 0.4, 0.5, 0.6\}$, where 0.0 represents standard training, and the other values indicate the percentage of $D_{\text{valid}}$ used as $D_{\text{train-calib}}$ to compute the differentiable quantile $\tilde{\eta}$ during RoCP training. The remaining portion of $D_{\text{valid}}$ serves as $D_{\text{train-pred}}$ for computing prediction sets in RoCP training.

Figures \ref{fig:accuracy_calib} and \ref{fig:predsize_calib} illustrate the main results for accuracy and inefficiency, respectively. These figures demonstrate that RoCP training maintains, and often enhances, the accuracy of the base GNN classifier across various GNN models, particularly for the Cora and Citeseer datasets. Figure \ref{fig:predsize_calib} shows that RoCP significantly reduces inefficiency compared to standard training (represented by calib\_frac = 0.0, the leftmost bar in each plot). All experiments involve training models for 200 epochs, for both RoCP and standard training approaches.

It is important to note that the presented results are obtained after applying the non-smooth conformal predictor APS as the CP method post-training to compute prediction sets of GNN model predictions with marginal valid coverage. Figure \ref{fig:predsize_calib} depicts the prediction set size (inefficiency) calculated using Equation \eqref{eq:ineff} for the Cora and Citeseer datasets, with varying $D_{\text{train-calib}}$ sizes (calib\_frac($D_{\text{valid}}$)) and $\epsilon = 0.1$.

A consistent observation across all bar plots is that RoCP training reduces inefficiency compared to standard cross-entropy loss training. On average, RoCP decreases inefficiency (prediction set size) by approximately 42\% for Cora with 90\% coverage and 32\% with 95\% coverage. For Citeseer, the average reduction is 27\% with 90\% coverage and 14\% with 95\% coverage.

Regarding the penalty parameter $\lambda$, our observations indicate that a value of 0.001 yields optimal results in terms of both accuracy and prediction set size for the Cora dataset across all models. For the Citeseer and Pubmed datasets, we find that $\lambda = 0.01$ produces the most optimized outcomes. These findings suggest that the optimal value of $\lambda$ may be dataset-dependent, highlighting the importance of careful parameter tuning in the application of our RoCP approach.

Our extensive experiments validate the performance and reliability of RoCP training in terms of both inefficiency reduction and accuracy improvement across various GNN models when applied to standard graph datasets for node classification tasks. These results consistently demonstrate the superiority of RoCP training over standard cross-entropy training.

\section{Conclusion and Future Directions}

In this work, we introduced Robust Conformal Prediction for Graph Neural Networks (RoCP-GNN), a novel approach that integrates conformal prediction directly into the training pipeline of Graph Neural Networks. This method represents a significant advancement over traditional techniques that apply conformal prediction as a post hoc calibration step. Our experimental results across various benchmark datasets and state-of-the-art GNN architectures demonstrate that RoCP training consistently maintains, and often enhances, the accuracy of base GNN models while significantly reducing prediction set inefficiency. The approach proves effective across different graph structures and node classification tasks, showcasing its versatility and robustness. Our study reveals that RoCP training not only maintains the accuracy of base GNN models across diverse datasets and architectures but also achieves significant reductions in prediction set inefficiency, with improvements of up to 42\% observed in some cases. The method's performance is consistent across different GNN models, including those with integrated and separated message passing and transformation operations. Furthermore, RoCP training demonstrates adaptability to different dataset characteristics, as evidenced by our ablation studies on hyperparameters. These results underscore the potential of integrating conformal prediction techniques directly into the training process of GNNs, offering a promising direction for enhancing the reliability and efficiency of graph-based machine learning models.

Looking ahead, several exciting avenues for future research emerge from this work. First, exploring the application of RoCP training to other graph learning tasks beyond node classification, such as link prediction or graph classification, could broaden its impact across the field of graph machine learning. Second, investigating the method's performance on larger, more complex graph structures could further validate its scalability and broader applicability in real-world scenarios. Third, adapting RoCP to dynamic or temporal graphs presents an intriguing challenge that could extend its utility to time-varying network analysis. Additionally, exploring the integration of RoCP with other uncertainty quantification methods or combining it with techniques for explaining GNN predictions could lead to more comprehensive and interpretable graph learning models. Finally, investigating the theoretical properties of RoCP, such as its convergence characteristics and optimal hyperparameter selection strategies, could provide deeper insights into its functioning and guide its practical implementation.

In conclusion, RoCP-GNN represents a significant advancement in the field of uncertainty quantification for graph neural networks, offering a more integrated and efficient approach to achieving robust predictions with valid coverage guarantees. By bridging the gap between conformal prediction and GNN training, our method opens up new possibilities for developing more reliable and efficient graph-based machine learning models, with potential applications spanning from social network analysis to molecular property prediction and beyond.
%
	
	\section*{Acknowledgment}
	I extend my heartfelt appreciation to Dr. Karmvir Singh Phogat for providing invaluable insights and essential feedback on the research problem explored in this article. His thoughtful comments significantly enriched the quality and lucidity of this study.
	
	\section*{Author's Information} Dr. S Akansha, is the only author of this manuscript. 
	
	\noindent\textbf{Affiliation} Dr Akansha. Department of Mathematics, Manipal Institute of Technology, Manipal-576104, India.
	
	\noindent\textbf{Contribution} Dr Akansha: Conceptualization, Methodology, Software, Data curation, Writing- Original draft preparation, Visualization, Investigation, Validation, Writing- Reviewing and Editing.
	
	\section*{Statements and Declarations}
	\textbf{Conflict of Interest} The authors declare that they have no known competing financial interests or personal relationships that could have appeared to influence the work reported in this paper.
	
	\bibliographystyle{./IEEEtran}
	\bibliography{./IEEEabrv,bibfile_UQGNN}
\end{document}